\def\year{2022}\relax
\documentclass[letterpaper,anonymous]{article} % DO NOT CHANGE THIS
\usepackage{aaai22}  % DO NOT CHANGE THIS
\usepackage{times}  % DO NOT CHANGE THIS
\usepackage{helvet}  % DO NOT CHANGE THIS
\usepackage{courier}  % DO NOT CHANGE THIS
\usepackage[hyphens]{url}  % DO NOT CHANGE THIS
\usepackage{graphicx} % DO NOT CHANGE THIS
\usepackage{natbib}  % DO NOT CHANGE THIS AND DO NOT ADD ANY OPTIONS TO IT
\usepackage{caption} % DO NOT CHANGE THIS AND DO NOT ADD ANY OPTIONS TO IT
\DeclareCaptionStyle{ruled}{labelfont=normalfont,labelsep=colon,strut=off} % DO NOT CHANGE THIS
\usepackage{algorithm}
\usepackage{algorithmic}

\usepackage{newfloat}
\usepackage{listings}
\usepackage{authblk}
\floatstyle{ruled}
\newfloat{listing}{tb}{lst}{}
\floatname{listing}{Listing}
\usepackage{amsmath}
\usepackage{amsfonts}
\usepackage{multicol}
\usepackage{multirow}
\usepackage{makecell}
\usepackage{amssymb}

\title{Improved Pillar with Fine-grained Feature for 3D Object Detection}
\author {
    Jiahui Fu\textsuperscript{\rm 1}$^{*}$,
    Guanghui Ren\textsuperscript{\rm 2}\thanks{These authors contributed to the work equally.},
    Yunpeng Chen\textsuperscript{\rm 2},
    Si Liu\textsuperscript{\rm 1}\thanks{The corresponding author is Si Liu.}
}
\affiliations {
    \textsuperscript{\rm 1}Beihang University \ \ \ 
    \textsuperscript{\rm 2}YiGo Robotics\\
    \{jiahuifu, liusi\}@buaa.edu.cn, \{sundrops.ren, qw.2080\}@gmail.com
}
\begin{document}

\maketitle

%-------------------------
\begin{abstract}

3D object detection with LiDAR point clouds plays an important role in autonomous driving perception module that requires high speed, stability and accuracy. However, the existing point-based methods are challenging to reach the speed requirements because of too many raw points, and the voxel-based methods are unable to ensure stable speed because of the 3D sparse convolution. In contrast, the 2D grid-based methods, such as PointPillar, can easily achieve a stable and efficient speed based on simple 2D convolution, but it is hard to get the competitive accuracy limited by the coarse-grained point clouds representation. So we propose an improved pillar with fine-grained feature based on PointPillar that can significantly improve detection accuracy. It consists of two modules, including height-aware sub-pillar and sparsity-based tiny-pillar, which get fine-grained representation respectively in the vertical and horizontal direction of 3D space. For height-aware sub-pillar, we introduce a height position encoding to keep height information of each sub-pillar during projecting to a 2D pseudo image. For sparsity-based tiny-pillar, we introduce sparsity-based CNN backbone stacked by dense feature and sparse attention module to extract feature with larger receptive field efficiently. Experimental results show that our proposed method significantly outperforms previous state-of-the-art 3D detection methods on the Waymo Open Dataset. The related code will be released to facilitate the academic and industrial study.

% while for Sparsity-based Tiny-pillar we introduce Sparsity-based  CNN  Backbone stacked by Dense  Feature  and  Sparse  Attention module to extract feature  with larger receptive field efficiently.

\end{abstract}

%-------------------------

%-------------------------
\section{Introduction}
3D object detection is an essential task for 3D scene perception and understanding, which aims at predicting 3D bounding boxes of objects based on sparse and irregular point clouds. 3D object detection has received increasing attention recently thanks to its wide applications in various fields such as autonomous driving and robotics.

%Conventional 3D object detection methods can mainly be sumarised into three types in terms of point cloud representations, \emph{i.e.,} the point-based methods, the 3D voxel-based methods and the 2D grid-based methods. The point-based methods utilize set abstraction layers to extract context features directly from raw point clouds, which can provide the most fine-grained 3D representations while requiring lots of computational costs to process large-scale point cloud data. The 3D voxel-based methods transform the point clouds to regular 3D voxel representations and extract features by 3D sparse convolutional neural networks. In this way, such methods can obtain detailed information of 3D objects by applying small voxel size and almost real-time execution speed with the high efficiency of 3D sparse convolution. The 2D grid-based methods transform the point clouds to a 2D pseudo image by projecting to birds-eye-view(BEV) or range-view(RV) and extract features by 2D convolutional neural networks. We argue that such methods can easily achieve a high speed with simple 2D convolution but lose some valid information during the projecting process. Thus, limited to the coarse-grained representation of point clouds, the 2D grid-based methods are hard to get accurate detection results as well as two other kinds of methods.

Conventional 3D object detection methods can mainly be summarized into three types in terms of point cloud representations, \emph{i.e.,} the point-based methods, the 3D voxel-based methods and the 2D grid-based methods. The point-based methods~\cite{DBLP:conf/nips/QiYSG17, DBLP:conf/cvpr/ShiWL19} utilize set abstraction layers to extract context features directly from raw point clouds. Large-scale raw point cloud data can provide the most fine-grained 3D representations but requires lots of computational costs to the process. The 3D voxel-based methods~\cite{zhou2018voxelnet, ye2020hvnet} transform the point clouds to regular 3D voxel representations and extract features by 3D sparse convolutional neural networks. Such methods can obtain detailed information of 3D objects by applying small voxel size but suffer from the unstable execution time of 3D sparse convolutions influenced by the point cloud density and hardware equipment. Finally, the 2D grid-based methods~\cite{lang2019pointpillars, meyer2019lasernet} transform the point clouds to a 2D pseudo image by projecting to birds-eye-view(BEV) or range-view(RV) and extract features by 2D convolutional neural networks. We argue that such methods can easily achieve a high speed with simple 2D convolution but lose some valid information during the projecting process. Thus, limited to the coarse-grained representation of point clouds, the 2D grid-based methods are hard to get accurate detection results as well as two other kinds of methods.

\begin{figure*}[ht]
\centering 
\includegraphics[width=\textwidth]{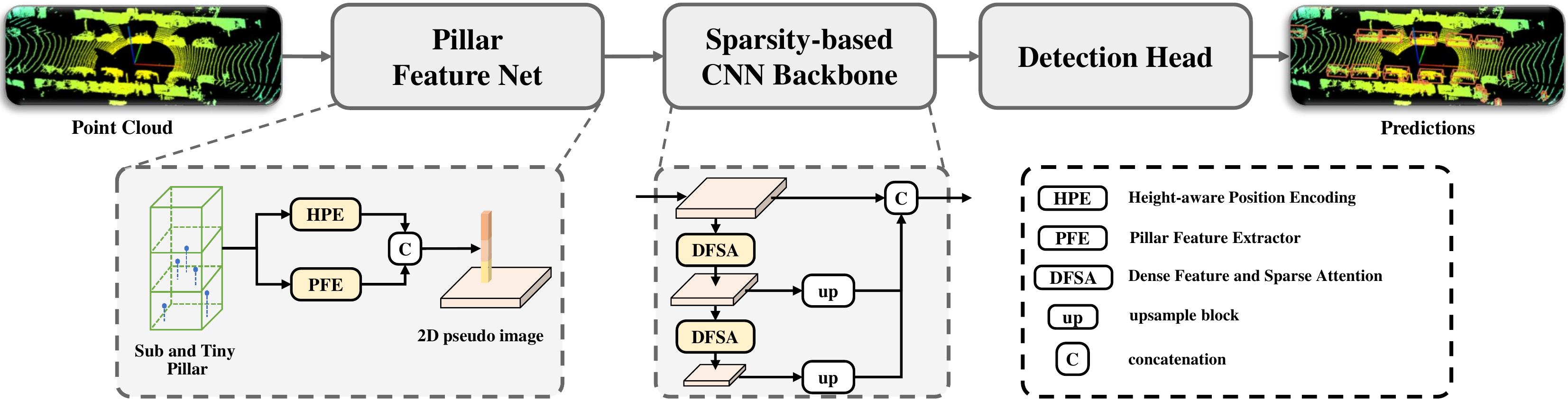}
\caption{The overview of our method for 3D object detection. The point clouds are first projected to a 2D pseudo image by pillar feature net. Then the image is fed into a 2D CNN backbone that is designed based on the sparsity of sub and tiny pillars for feature extraction. Finally, the detection head predicts the 3D bounding box of objects based on extracted features.}
\label{fig:method-0}
\end{figure*}

%Considering the requirements of the actual real-time system, heavy MLPs architectures with expensive computational cost in the point-based methods leading to unacceptable slow execution speed, 3D sparse convolutions in the 3D voxel-based methods take unstable execution time influenced by the point cloud density and hardware equipment. While the 2D grid-based methods can get high and stable execution speed with simple 2D convolutions, it will have high actual application value if can reach a competitive detection accuracy. 
To this end, we propose an improved Pillar with Fine-grained Feature for 3D object detection as shown in Figure~\ref{fig:method-0} based on a classic 2D grid-based method PointPillar~\cite{lang2019pointpillars}. We argue that our proposed method makes the pillar-based representation of point clouds better in two aspects: 
\textit{First}, we notice that the point cloud is unevenly distributed in the height dimension, making the representation of the origin pillar is mainly dominated by the area where the points are concentrated. To solve it, we introduce a Sub-Pillar to get fine-grained representation in the vertical direction by splitting multi sub-pillars in the height dimension to force the retention of information of other areas. 
\textit{Second}, we consider that the 2D grid size is too larger compared with the 3D voxel size, which makes it difficult to learn the detailed feature representation of small objects. Thus, we introduce a Tiny-Pillar to get fine-grained representation in the horizontal direction by applying a smaller grid size in the 2D pseudo image.

Applying the above simple strategies can improve the accuracy of detection but also bring some problems: \textit{First}, the different height position information of each sub-pillar is lost during projecting to a 2D pseudo image. Therefore, we propose a Height-aware Sub-Pillar~(HS-Pillar), which uses height-aware position encoding to represent the height position information for each sub-pillar explicitly. 
\textit{Second}, simply reducing the grid size causes more execution time cost and smaller receptive field. So we propose a Sparsity-based Tiny-Pillar~(ST-Pillar), which uses a sparsity-based CNN backbone for more efficient feature extraction and larger receptive field from a tiny-pillar feature map. The sparsity-based CNN backbone is stacked by the Dense Feature and Sparse Attention~(DFSA) module, representing the distribution of objects on sparse large-scale features and extracting fine-grained local object features on dense small-scale features.

To be summarized, our main contributions are as follow:
\begin{itemize}
\item We propose an HS-Pillar to get fine-grained representation in the vertical direction. Specifically, we split multi sub-pillars in the height dimension and propose a height-aware position encoding to keep height position information.
\item We propose an ST-Pillar to get fine-grained representation in the horizontal direction. In this way, we reduce the grid size in a 2D pseudo image and propose a sparsity-based CNN backbone based on the DFSA module to extract features efficiently.
\item We demonstrate that our approach could achieve new state-of-the-art results on the Waymo Open dataset for mAP/mAPH in LEVEL 1 and LEVEL 2 difficulties.
\end{itemize}

%-------------------------

%-------------------------
\section{Related Work}
\subsection{3D Object Detection with Point-based Methods}
The point-based methods are mostly based on the PointNet++~\cite{DBLP:conf/nips/QiYSG17}, especially the set abstraction operation, enabling flexible receptive fields for point cloud feature learning. PointRCNN~\cite{DBLP:conf/cvpr/ShiWL19} generates 3D proposals directly from the whole point clouds. STD~\cite{DBLP:conf/iccv/YangS0SJ19} proposes a sparse-to-dense strategy to refine proposals. VoteNet~\cite{DBLP:conf/iccv/QiLHG19} proposes the hough voting strategy to get the key points of each object. SD-SSD~\cite{DBLP:conf/cvpr/YangS0J20} proposes a sampling strategy based on the feature distance to sample more object points. Pointformer~\cite{DBLP:journals/corr/abs-2012-11409} uses a transform structure instead of MLPs to extract features from point sets. These methods can get accuracy detection results, but all rely on large-scale point cloud data and the heavy MLPs architecture of set abstraction operation, so their execution speed is difficult to meet real-time requirements.
\subsection{3D Object Detection with 3D Voxel-based Methods}
The 3D-voxel based methods mostly divide the point clouds into 3D voxels to be processed by 3D sparse CNN~\cite{yan2018second}. VoxelNet~\cite{zhou2018voxelnet} applies MLPs to points in each voxel and gets 3D feature maps. HVNet~\cite{ye2020hvnet} aggregates multi-level features by applying multi-scale voxelization to the point cloud. \cite{shi2020pv, miao2021pvgnet} introduces features of key points and fuses them with voxel features to get more accurate location information. The key of these methods that can achieve a balance of accuracy and speed is 3D sparse convolution. But it makes the execution time of these methods dependant on the density of the point cloud, which leads to instability in the practical application with various environments. Also, if more data sources are available to get better performance, such as multiple LiDAR sensors or multi frames fusion, the execution time of these methods will increase a lot due to processing denser point clouds. Besides, the hardware platform has worse support for 3D sparse convolution than simple 2D convolution, which creates certain obstacles for applying 3D voxel-based methods in the real world.
\subsection{3D Object Detection with 2D Grid-based Methods}
The 2D grid-based methods mostly project the point clouds to 2D birds-eye-view or range-view to be processed by 2D CNN. PointPillars~\cite{lang2019pointpillars} learns a representation of point clouds organized in vertical columns and gets a 2D feature in birds-eye-view. LaserNet~\cite{meyer2019lasernet} uses a fully convolutional network on range-view images to predict a distribution over 3D boxes for each point. RSN~\cite{sun2021rsn} predicts foreground points from range-view images and applies sparse convolutions on the selected foreground points to detect objects. CVC-Net~\cite{chen2020every} proposes a pair of cross-view transformers to transform the feature maps between birds-eye-view and range-view. Operating in the range view involves well-known disadvantages for object detection, including occlusion and scale variation, so we prefer to use the birds-eye-view feature to get more accurate detection results. However, most of the existing birds-eye-view representations for detection, represented by pillar-based methods, are not as accurate as the other two kinds of methods, especially for small objects. In our work, we analyze the bottleneck of previous pillar-based methods and find that they are mainly limited by coarse-grained representation, which specifically reflects in the lack of distribution information of vertical direction during projecting into a 2D image and detailed information of horizontal direction during splitting into large grids. So we propose a more fine-grained pillar for the above two bottlenecks.

%-------------------------

%-------------------------

\begin{figure}[t]
\centering 
\includegraphics[width=0.45\textwidth]{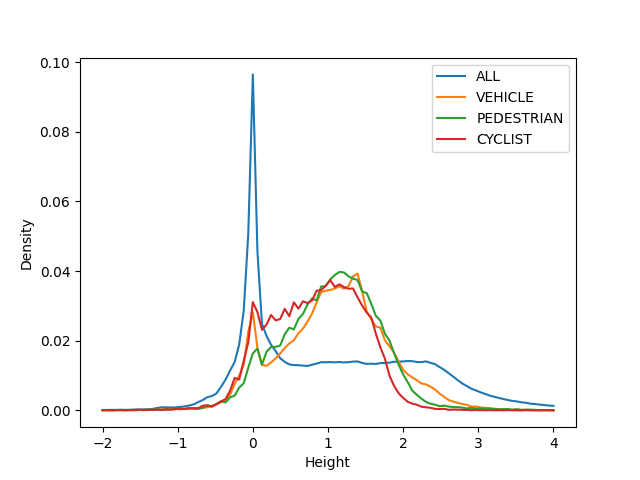}
\caption{A study on the distribution of point clouds in the height dimension on the Waymo Open Dataset. Here we randomly sample 100 scenes and respectively counted the height distribution of the point clouds in the whole scene and boxes of three categories.}
\label{fig:method-1}
\end{figure}

\section{Method}
% TODO: a summary
To improve the point cloud representations based on pillars, we propose Height-aware Sub-pillar and Sparsity-based Tiny-pillar to get fine-grained features respectively in the vertical and horizontal direction. As shown in Figure~\ref{fig:method-0}, our method consists of three main steps. First, we project the point clouds into a sub and tiny pillar to get the fine-grained 2D pseudo image. Second, we use a CNN backbone consisting of the DFSA module designed based on the sparsity of pseudo images to extract features, which contain position information of objects from large-scale feature maps and shape information of objects from small-scale feature maps. Third, the extracted features are feed into the detection head to predict the size and location of 3D bounding boxes.

\begin{figure}[t]
\centering 
\includegraphics[width=0.45\textwidth]{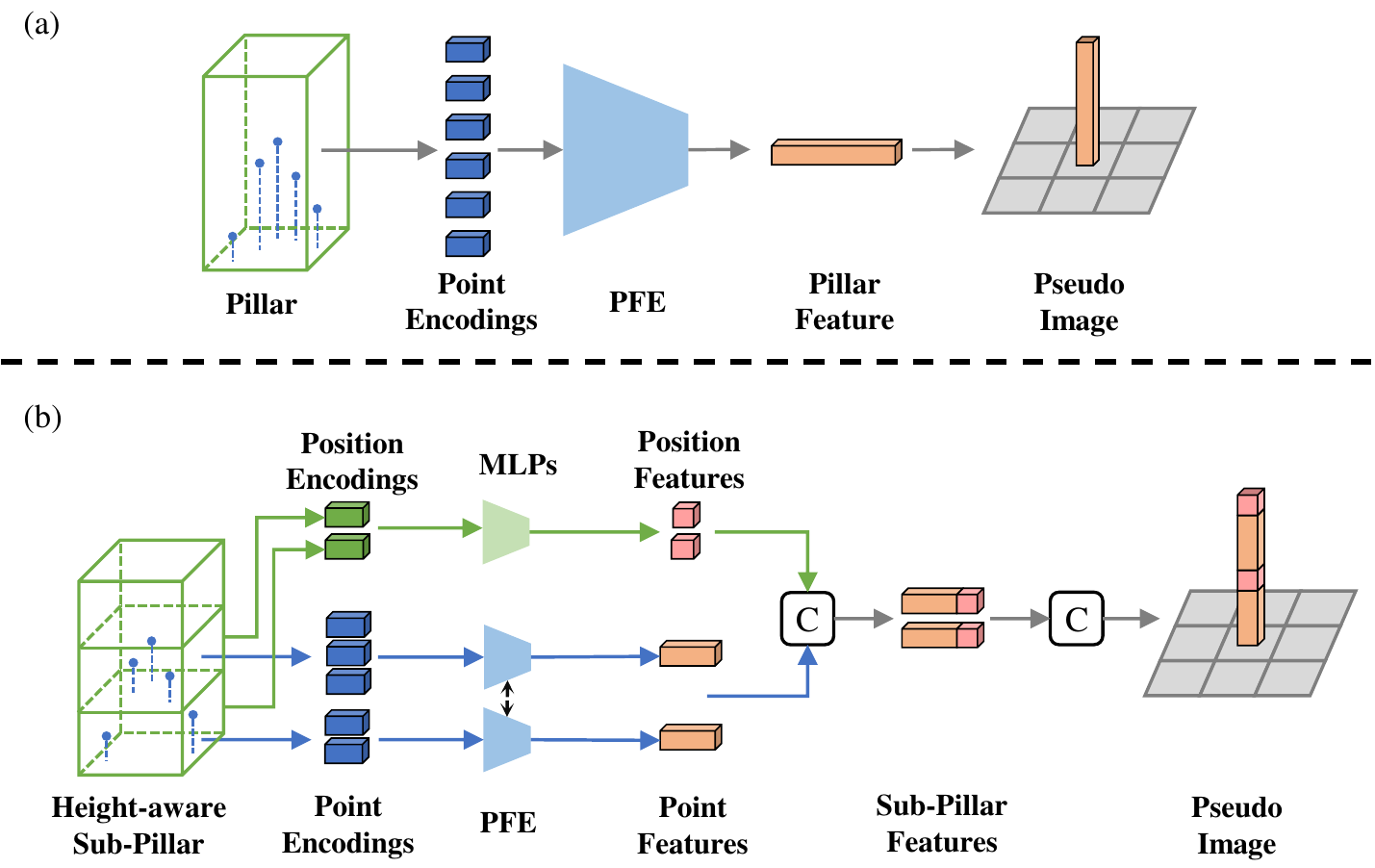}
\caption{A comparison between pointpillar $(a)$ and our height-aware sub-pillar $(b)$, our method divides the pillar into multiple sub-pillars in the height dimension and extracts the feature of the points separately, we also add a height-aware position feature to each sub-pillar.}
\label{fig:method-2}
\end{figure}

\subsection{Height-aware Sub-pillar}
\noindent\textbf{Sub-pillar.}~The 2D grid-based methods project the point cloud to 2D pseudo images by extracting the feature from pillars, which contains all points of the entire range in the height dimension. But the position distribution of points in the height dimension is extremely uneven. As the Figure~\ref{fig:method-1} shows, the height distribution of the whole scene points is highly concentrated near $0m$, and the points of objects are mostly distributed from $0m$ to $2m$. So the learned global feature of each pillar with the extremely uneven distribution by limited linear layers is always dominated by the concentrated area containing most of the points. However, the information of a small number of points in other areas is critical to accurately predict the boundary of the object, which are almost ignored in the extraction process of the entire pillar feature due to a very small proportion. So we need more fine-grained features to represent the point cloud distribution at various heights.

To enforce the features projected to the 2D pseudo image containing each area's information in the height dimension, we propose a sub-pillar to get height dimensionally local features. As the Figure~\ref{fig:method-2} shows, we split each pillar into $N_h$ sub-pillars and use two VFE layers~\cite{zhou2018voxelnet} as Pillar Feature Extractor~(PFE) to extract point features from each sub-pillar. Then, we can simply concatenate all the sub-pillar features together as the feature of each position in the 2D pseudo image. Specifically, The input point encoding to PFE is raw data of points augmented with $(x_c, y_c, z_c, x_p, y_p, z_p)$ where the $c$ subscript denotes distance to the mean of all points in the sub-pillar and the $p$ subscript denotes the offset from the sub-pillar $(x, y, z)$ center, which follow the setting of PointPillar~\cite{lang2019pointpillars}.

\begin{figure}[t]
\centering 
\includegraphics[width=0.45\textwidth]{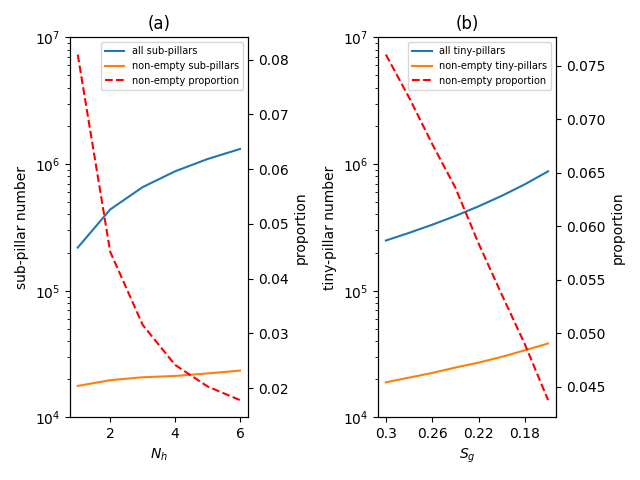}
\caption{A study on the trend of pillars number with splitting number $N_h$and the grid size $S_g$. Here we randomly sample 100 scenes for statistics, it can be seen that as $N_h$ increases and $S_g$ decreases, the all number of pillars increases significantly, but the number of not empty pillars increases relatively slowly and the proportion of them rapidly decreases. To be noticed that the left ordinate here uses a logarithmic scale.}
\label{fig:method-3}
\end{figure}

To be noticed that we only extract the features of the sub-pillars with points inside, and the position distribution of points in height dimension is concentrated locally, so the sub-pillar only brings a slight increase in calculation during pillar feature extraction. As Figure~\ref{fig:method-3} $(a)$ shows that as more sub-pillars are split in each position, the increase in the number of not empty sub-pillars is very limited.

\begin{figure}[t]
\centering 
\includegraphics[width=0.45\textwidth]{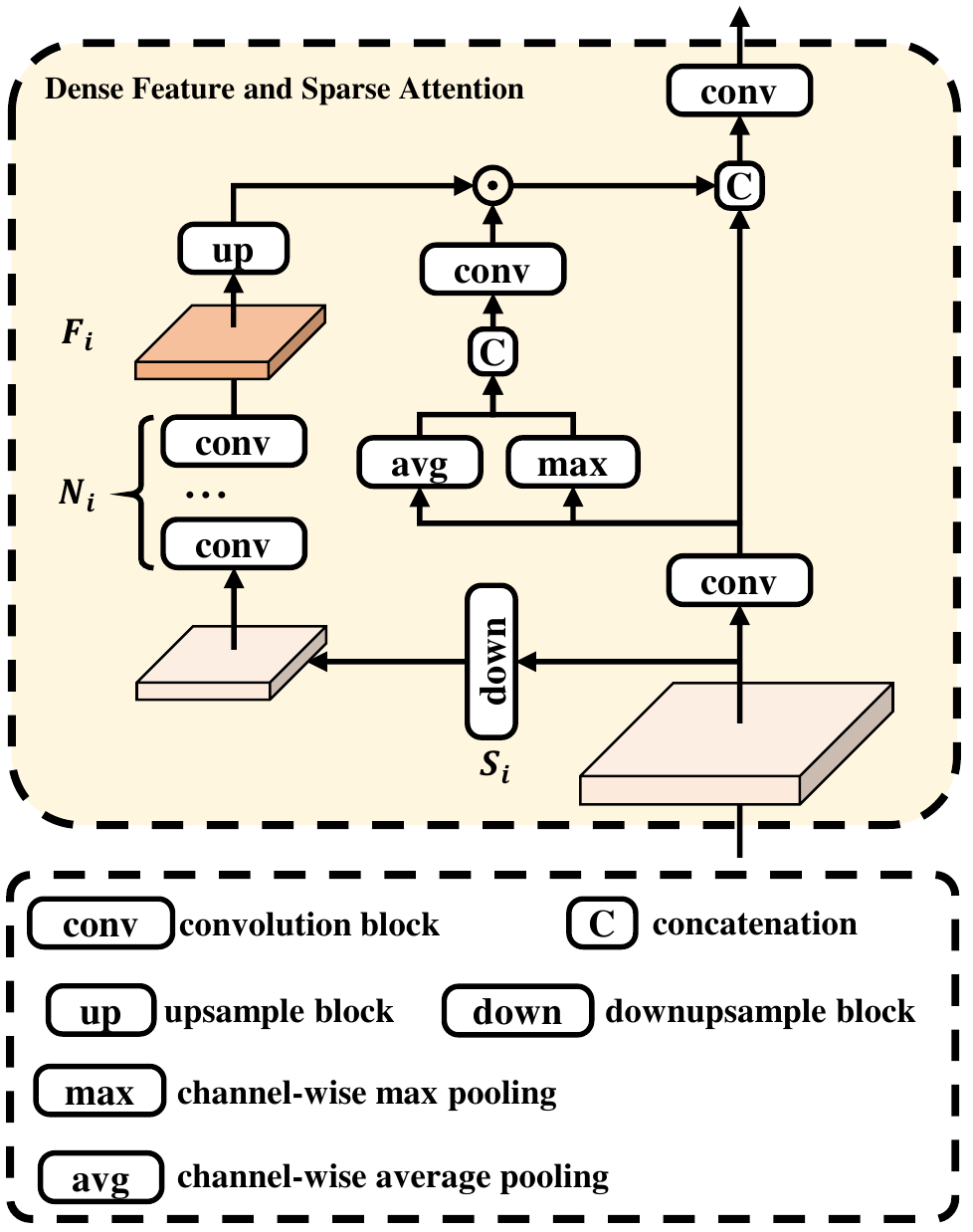}
\caption{Structure details of DFSA module. It contains a large-scale feature branch, a sparse attention branch and multi small-scale feature branches.}
\label{fig:method-4}
\end{figure}

\noindent\textbf{Height-aware Position Encoding.}~Simply concatenating all the sub-pillar point features as the global feature causes one problem: height position information of each sub-pillar lost in the 2D pseudo image feature. So we propose a Height-aware Position Encoding~(HPE) to explicitly represent each sub-pillar's height position information. As the Figure~\ref{fig:method-2} shows that we add a position encoding for each sub-pillar based on the mean of all points height $z_m$ and height of center $z_p$, which can be formulated as:
\begin{equation}
\begin{aligned}
    P(z) &= \{(sin(2^i \pi z), cos(2^i \pi z) |_{i=0}^{L-1}\}\\
    z &= (z_m, z_p)
\end{aligned}
\end{equation}
Furthermore, the features of each sub-pillar are composed of point feature and position feature by concatenation, and the 2D pseudo image feature is composed of sub-pillar features at the same position.

\subsection{Sparsity-based Tiny-pillar}
\noindent\textbf{Tiny-pillar.}~Thanks to the efficient 3D sparse convolutions, the 3D voxel-based methods can apply a small voxel size compared with 2D grid size to get more fine-grained features. For example, in Waymo Open Dataset~\cite{sun2020scalability}, the 3D voxel size is usually set to $(0.1m, 0.1m, 0.15m)$ ~\cite{shi2020pv}, while the 2D grid size is $(0.32m, 0.32m, 6m)$~\cite{yin2021center}, so one pillar contains almost 400 voxels in space, which limits the performance of 2D grid-based methods especially in small object detection. So we propose a tiny-pillar to get spatially fine-grained features of a 2D pseudo image by reducing the size of the grid significantly to the half of original setting.

\noindent\textbf{Sparsity-based CNN Backbone.}~Simply reducing the size of the grid causes two serious problems. Firstly, the size of the 2D pseudo image increases, further causes a significant execution time cost at the stage of the 2D CNN backbone. Secondly, the absolute size of the receptive field in the 3D space reduces because the 3D space represented by each tiny-pillar has become smaller.

Therefore, to extract fine-grained features and keep less execution time and large receptive field simultaneously, we proposed a Sparsity-based CNN Backbone~(SCB) stacked by Dense Feature and Sparse Attention~(DFSA) module. The DFSA module is designed based on the fact that most of the increased tiny-pillars due to reducing the grid size $S_g$ are empty. As the Figure~\ref{fig:method-3} $(b)$ shows that the proportion of not empty tiny-pillars decreases significantly as the grid size decreases. Thus, applying too many convolution blocks directly on the original large-size pseudo image is inefficient and unnecessary. Instead, we can represent the distribution of objects on sparse large-scale features to predict more accurate object centers and extract fine-grained local object features on dense small-scale features to predict more accurate object boundaries at the same time.

As shown in Figure~\ref{fig:method-4}, in each DFSA module of SCB, the input sparse large-scale feature map only pass one strided convolution block to keep position information of objects. Then we apply max pooling and average pooling layers to the large-scale feature along the channel dimension. Next, we concatenate the pooled features and feed them to a $7\times7$ convolution layer, followed by applying a sigmoid function, to produce the spatial attention map. Meanwhile, we respectively downsample the large-scale input feature map by $\{S_i\}_{1}^{n}$ times and apply $\{N_i\}_1^n$ convolution blocks to get a series of dense small-scale feature maps $\{F_i\}_1^n$. The number of convolution blocks $\{N_i\}_1^n$ applying to the downsampled features increases as the scale decreases. This simple strategy can make the feature extraction more efficient and the receptive field larger by more convolution blocks on smaller feature maps. Finally, the downsampled feature maps $\{F_i\}_1^n$ with fine-grained local object features are upsampled to output feature map size guided by the above spatial attention map. The output of each DFSA module is features from all branches concatenated and fused by a $1\times1$ convolution block. The output of SCB is features from all modules concatenated based on input feature map size.

%-------------------------

%-------------------------
\begin{table*}[ht]
    \centering
    \resizebox{\textwidth}{!}{
    \begin{tabular}{c|c|cc|cc|cc|cc|cc|cc}
        \hline
        \multirow{2}*{Type} & \multirow{2}*{Method} & \multicolumn{2}{|c|}{Veh. (LEVEL 1)} & \multicolumn{2}{|c|}{Veh. (LEVEL 2)} & \multicolumn{2}{|c|}{Ped. (LEVEL 1)} & \multicolumn{2}{|c|}{Ped. (LEVEL 2)} & \multicolumn{2}{|c|}{Cyc. (LEVEL 1)} & \multicolumn{2}{|c}{Cyc. (LEVEL 2)}\\
        ~ & ~  & mAP & mAPH & mAP & mAPH & mAP & mAPH & mAP & mAPH & mAP & mAPH & mAP & mAPH \\
        \hline
        \multirow{1}*{\makecell[c]{point-based}} & StarNet~\cite{ngiam2019starnet}& 53.70 & - & - & - & 66.80 & - & - & - & - & - & - & -\\
        \hline
        \multirow{5}*{\makecell[c]{3D \\ voxel-based}} & SECOND~\cite{yan2018second} & 72.27 & 71.69 & 63.85 & 63.33 & 68.70 & 58.18 & 60.72 & 51.31 & 60.62 & 59.28 & 58.34 & 57.05\\
        ~ & Part-A2-Net~\cite{shi2020points} & 74.82 & 74.32 & 65.88 & 65.42 & 71.76 & 63.64 & 62.53 & 55.30 & 67.35 & 66.15 & 65.05 & 63.89 \\
        ~ & PV-RCNN~\cite{shi2020pv} & 77.51 & 76.89 & 68.98 & 68.41 & 75.01 & 65.65 & 66.04 & 57.61 & 67.81 & 66.35 & 65.39 & 63.98\\
        ~ & PV-RCNN++~\cite{shi2021pv} & 78.79 & 78.21 & 70.26 & 69.71 & 76.67 & 67.15 & 68.51 & 59.72 & 68.98 & 67.63 & 66.48 & 65.17 \\
        ~ & CenterPoint-Voxel~\cite{yin2021center} & 76.70 & 76.20 & 68.80 & 68.30 & 79.00 & 72.90 & 71.00 & 65.30 & - & - & - &\\
        \hline
        \multirow{7}*{\makecell[c]{2D \\ grid-based}} & PointPillar~\cite{lang2019pointpillars} & 56.62 & - & - & - & 59.25 & - & - & - & - & - & - & - \\
        ~ & MVF~\cite{zhou2020end} & 62.93 & - & - & - & 65.33 & - & - & - & - & - & - & - \\
        ~ & Pillar-based~\cite{wang2020pillar} & 69.80 & - & - & - & 72.51 & - & - & - & - & - & - & - \\
        ~ & CenterPoint-Pillar~\cite{yin2021center} & 76.10 & 75.50 & 68.00 & 67.50 & 76.10 & 65.10 & 68.10 & 57.90 & - & - & - & -\\
        ~ & {\dag}CenterPoint-Pillar & 74.38 & 73.82 & 66.09 & 65.58 & 72.35 & 63.52 & 65.50 & 56.39 & 63.81 & 62.32 & 61.48 & 60.04 \\
        ~ & \textbf{HS-Pillar(Ours)} & 75.02 & 74.49 & 66.73 & 66.24 & 75.26 & 66.70 & 67.37 & 59.46 & 67.61 & 66.13 & 65.13 & 63.70 \\
        ~ & \textbf{ST-Pillar(Ours)} & 77.47 & 77.00 & 69.42 & 68.98 & 79.24 & 71.99 & 71.05 & 64.27 & 69.61 & 68.40 & 66.85 & 65.69 \\
        ~ & \textbf{HS\&ST-Pillar(Ours)} & \textbf{79.05} & \textbf{78.39} & \textbf{70.29} & \textbf{70.25} & \textbf{80.13} & \textbf{73.15} & \textbf{72.24} & \textbf{65.67} & \textbf{71.49} & \textbf{70.39} & \textbf{68.82} & \textbf{67.76} \\
        \hline
    \end{tabular}
    }
    \caption{Performance comparison on the Waymo Open Dataset with 202 validation sequences. \dag: re-implemented one-stage model by ourselves with their open source code for converting from two categories to three categories.}
    \label{tab:exp-1}
\end{table*}

% \begin{table*}[t]
%     \centering
%     \resizebox{0.85\textwidth}{!}{
%     \begin{tabular}{c|ccc|ccc|ccc|ccc}
%         \hline
%         \multirow{2}*{Method} & \multicolumn{3}{|c|}{Car} & \multicolumn{3}{|c|}{Pedestrian} & \multicolumn{3}{|c}{Cyclist} & \multicolumn{3}{|c}{Mean} \\
%         ~ & Easy & Mod & Hard & Easy & Mod & Hard & Easy & Mod & Hard & Easy & Mod & Hard \\
%         \hline
%         {\dag}CenterPoint-Pillar & 83.25 & 73.86 & 69.93 & 40.80 & 36.73 & 34.01 & 81.88 & 61.88 & 57.57 & 68.64 & 57.49 & 53.84 \\
%         \textbf{HS-Pillar(Ours)} & 86.05 & 75.66 & 72.79 & 44.60 & 40.76 & 37.53 & 80.92 & \textbf{62.89} & \textbf{58.95} & 70.52 & 59.77 & 56.42 \\
%         \textbf{ST-Pillar(Ours)} & \textbf{88.02} & 78.57 & 76.18 & 51.83 & 47.11 & 43.53 & \textbf{87.69} & 61.99 & 58.17 & 75.85 & 62.56 & 59.29 \\
%         \textbf{HS\&ST-Pillar(Ours)} & 87.98 & \textbf{78.89} & \textbf{76.24} & \textbf{55.46} & \textbf{50.09} & \textbf{45.49} & 85.03 & 60.46 & 57.25 & \textbf{76.16} & \textbf{63.15} & \textbf{59.66} \\
%         \hline
%     \end{tabular}
%     }
%     \caption{Performance comparison on the KITTI Dataset with 3769 val samples. \dag: re-implemented one-stage model by ourselves with their open source code.}
%     \label{tab:exp-2}
% \end{table*}

\begin{table}[th]
    \centering
    \resizebox{0.45\textwidth}{!}{
    \begin{tabular}{c|ccc|ccc}
        \hline
        \multirow{2}*{Method} & \multicolumn{3}{|c|}{Car} & \multicolumn{3}{|c}{Pedestrian} \\
        ~ & Easy & Mod & Hard & Easy & Mod & Hard \\
        \hline
        {\dag}CenterPoint-Pillar & 83.25 & 73.86 & 69.93 & 40.80 & 36.73 & 34.01 \\
        \textbf{HS-Pillar(Ours)} & 86.05 & 75.66 & 72.79 & 44.60 & 40.76 & 37.53 \\
        \textbf{ST-Pillar(Ours)} & \textbf{88.02} & 78.57 & 76.18 & 51.83 & 47.11 & 43.53 \\
        \textbf{HS\&ST-Pillar(Ours)} & 87.98 & \textbf{78.89} & \textbf{76.24} & \textbf{55.46} & \textbf{50.09} & \textbf{45.49} \\
        \hline
    \end{tabular}
    }
    \caption{Performance comparison on the KITTI Dataset with 3769 val samples. \dag: re-implemented one-stage model by ourselves with their open source code.}
    \label{tab:exp-2}
\end{table}

\begin{table}[]
    \centering
    \resizebox{0.45\textwidth}{!}{
    \begin{tabular}{c|ccc|ccc}
        \hline
        \multirow{2}*{Method} & \multicolumn{3}{|c|}{Pedestrian 3D mAP (LEVEL 1)} & \multicolumn{3}{|c}{Cyclist 3D mAP (LEVEL 1)} \\
        ~ & 0-30m & 30-50m & 50m-inf & 0-30m & 30-50m & 50m-inf \\
        \hline
        SECOND & 74.39 & 67.24 & 56.71 & 73.33 & 55.51 & 41.98 \\
        PointPillar & 67.99 & 57.01 & 41.29 & - & - & - \\
        MVF & 72.51 & 63.35 & 50.62 & - & - & - \\
        Pillar-based & 79.34 & 72.14 & 56.77 & - & - & - \\
        Part-A2-Net & 81.87 & 73.65 & 62.34 & \textbf{80.87} & 62.57 & 45.40 \\
        PV-RCNN++ & 82.41 & 75.42 & 66.01 & 80.76 & 63.10 & 47.40 \\
        {\dag}CenterPoint-Pillar & 73.88 & 74.12 & 65.41 & 70.78 & 61.68 & 49.14 \\
        \hline
        \textbf{HS-Pillar(Ours)} & 80.25 & 74.58 & 65.47 & 75.44 & 64.49 & 51.33 \\
        \textbf{ST-Pillar(Ours)} & 83.75 & 78.45 & 69.28 & 78.29 & 64.38 & 53.67 \\
        \textbf{HS\&ST-Pillar(Ours)} & \textbf{84.99} & \textbf{78.84} & \textbf{70.17} & 79.30 & \textbf{67.44} & \textbf{55.14} \\
        \hline
    \end{tabular}
    }
    \caption{Performance at different distance ranges comparison on the Waymo Open Dataset with 202 validation sequences for the pedestrian and cyclist detection. \dag: re-implemented one-stage model by ourselves with their open source code.}
    \label{tab:exp-3}
\end{table}

\section{Experiments}
\subsection{Datasets}
We evaluate our Fine-grained Pillar on Waymo Open Dataset and KITTI Dataset,

\noindent\textbf{Waymo Open Dataset.}~Waymo Open Dataset~\cite{sun2020scalability} contains $798$ training sequences and $202$ validation sequences for vehicle, pedestrian and cyclist. The official evaluation toolkit also provides a performance breakdown for two difficulty levels: LEVEL 1 for boxes with more than five Lidar points, and LEVEL 2 for boxes with at least one Lidar point. The official 3D detection evaluation metrics include the standard 3D bounding box mean average precision (mAP) and mAP weighted by heading accuracy (mAPH). The mAP and mAPH are based on an IoU threshold of $0.7$ for vehicles and $0.5$ for pedestrians and cyclists.

\noindent\textbf{KITTI Dataset.}~KITTI Dataset~\cite{geiger2013vision}contains $7481$ training samples and $7518$ test samples, where the training samples are generally divided into $3712$ samples of train split and $3769$ samples of val split. Object instances across different classes are further classified into easy, moderate and hard splits, depending on the object size, the degree of occlusion and the maximum truncation level. The official 3D detection evaluation metric is mean average precision (mAP) with $40$ recall positions~\cite{simonelli2019disentangling}, which is based on an IoU threshold of $0.7$ for vehicles and $0.5$ for pedestrians and cyclists.
\subsection{Implementation Details}
\noindent\textbf{Hyper-parameters.}~ The number $N_h$ of split pillars in the height dimension is defaults to $6$ for the Waymo Open Dataset and $4$ for the KITTI Dataset, we also reduce the grid size $S_g$ to half of the original settings from the prior works~\cite{shi2020pv, lang2019pointpillars}, which means reducing $(0.32m, 0.32m)$ to $(0.16m, 0.16m)$ for the Waymo Open Dataset and $(0.16m, 0.16m)$ to $(0.08m, 0.08m)$ for the KITTI Dataset. The network is trained for 24 epochs with the batch size 16 for the Waymo Open Dataset and 80 epochs with the batch size 16 for the KITTI Dataset.

\noindent\textbf{Detection heads.}~ Following the prior work~\cite{yin2021center, wu2020iou}, we use center heatmap head, regression heads for a center location refinement $o \in \mathbb{R}$, height-above-ground $h_g \in \mathbb{R}$, the 3D size $s \in \mathbb{R}^3$, a yaw rotation angle $(sin(\alpha), cos(\alpha)) \in \mathbb{R}^2$ and a IoU prediction between object detections and corresponding ground truth boxes. At training time, the center heatmap head is supervised using a focal loss and only ground truth centers are supervised using an L1 regression loss for each regression head. At inference time, we extract all properties by indexing into dense regression head outputs at each object’s peak location and apply an IoU-Aware confidence rectification to each object score from center heatmap.

\subsection{Quantitative Analysis}
\noindent\textbf{Comparison with state-of-the-art methods on Waymo Open Dataset.}~As shown in Table \ref{tab:exp-1}, our methods can achieve remarkably better mAP/mAPH on all difficulty levels for the detection of all three categories of Waymo Open Dataset. Our proposed fine-grained pillar called HS\&ST-Pillar outperforms previous state-of-the-art method PV-RCNN++~\cite{shi2021pv} for all three categories significantly with $+0.54\%$, $+5.95\%$ and $+2.59\%$ mAPH LEVEL 2 gain respectively, also for previous state-of-the-art method CenterPoint-Voxel~\cite{yin2021center} only is trained and test for vehicle and pedestrian categories, our method for all categories can keep $+1.95\%$ and $+0.37\%$ mAPH LEVEL 2 gain respectively.

Furthermore, for a fair comparison, we re-implement one-stage CenterPoint-Pillar~\cite{yin2021center} with their open source code on three categories instead of only two categories as our baseline model. Our proposed methods can achieve significant and stable improvements for all three categories based on the baseline model, especially for small objects such as pedestrian and cyclist, which is because that point clouds of small objects have more obvious uneven distribution in the height dimension and rely more on smaller grid size to get fine-grained feature. Specifically, our proposed HS-Pillar can get $+0.66\%$, $+3.07\%$ and $+3.66\%$ mAPH improvement on LEVEL 2 respectively for three categories compared with the baseline model, which validates that our HS-Pillar can improve the feature extracted from the unevenly distributed point clouds. Our proposed ST-Pillar can achieve a more significant improvement with $+3.40\%$, $+4.81\%$ and $+5.65\%$ mAPH LEVEL 2 gain compared with the baseline model, which validates the fine-grained features based on the small grid.

To better demonstrate the performance at different distance ranges, we also present the distance-based detection performance in Table~\ref{tab:exp-3} for pedestrian and cyclist respectively. We can see that our HS\&ST-Pillar achieves best performance on all distance ranges of pedestrian 3D detection, and on more distance ranges for cyclist 3D detection. It is worth noting that Table~\ref{tab:exp-3} shows that our proposed method achieves much better performance than previous methods in terms of the furthest area ($>$50m), where the small objects are difficult to be accurately detected. Our HS\&ST-Pillar outperforms previous state-of-the-art method PV-RCNN++ with a $+4.16\%$ 3D mAP gain for pedestrian detection and CenterPoint-Pillar with a $+6.00\%$ 3D mAP gain for cyclist detection in area 50m-inf. These significant improvements benefit from the better feature representation of fine-grained pillar.

\noindent\textbf{Comparison with baseline method on KITTI Dataset.}~We also verify the improvements of our methods on the KITTI dataset. As shown in Table~\ref{tab:exp-2}, For large object such as car, our proposed HS\&ST-Pillar can achieve a significant improvement with $+4.73\%$, $+5.03\%$ and $+6.31\%$ mAP gain for three difficulty level compared with the baseline model, which shows our fine-grained feature can benefit hard object detection with serious occlusion and truncation. For small object such as pedestrian, our proposed HS\&ST-Pillar can achieve a much more significant improvement with $+14.66\%$, $+13.36\%$ and $+11.48\%$ mAP gain.

\subsection{Ablation Studies}
For efficiently conducting the ablation experiments on the Waymo Open dataset, we generate a small representative training set by uniformly sampling $20\%$ frames from the training set and validation set, all results are evaluated with the official evaluation tool of Waymo dataset.

\begin{table}[ht]
    \centering
    \resizebox{0.45\textwidth}{!}{
    \begin{tabular}{ccccc|c|c|c|c}
        \hline
        Sub-Pillar & PE & Tiny-Pillar & DF & SA & Veh. & Ped. &  Cyc. & Mean\\
        \hline
        & & & & & 60.32 & 49.22 & 51.58 & 53.71\\
        \hline
        \checkmark & & & & & 60.13 & 52.32 & 54.91 & 55.79\\
        \checkmark & \checkmark & & & & 61.35 & 53.50 & 56.58 & 57.14\\
        \hline
        & & \checkmark & & & 63.11 & 56.89 & 61.74 & 60.58\\
        & & \checkmark & \checkmark & & 67.33 & 59.24 & 65.06 & 63.88\\
        & & \checkmark & \checkmark & \checkmark & 67.22 & 59.34 & 65.82 & 64.12\\
        \checkmark & \checkmark & \checkmark & \checkmark & \checkmark & 67.32 & 60.49 & 66.04 & 64.62\\
        \hline
    \end{tabular}
    }
    \caption{Effects of our HS-Pillar with Position Encoding, ST-Pillar with Dense Feature and Sparse Attention. We compare on $20\%$ Waymo validation by per-class and average 3D LEVEL 2 mAPH.}
    \label{tab:exp-4}
\end{table}

\begin{figure*}[ht]
    \centering
    \includegraphics[width=\textwidth]{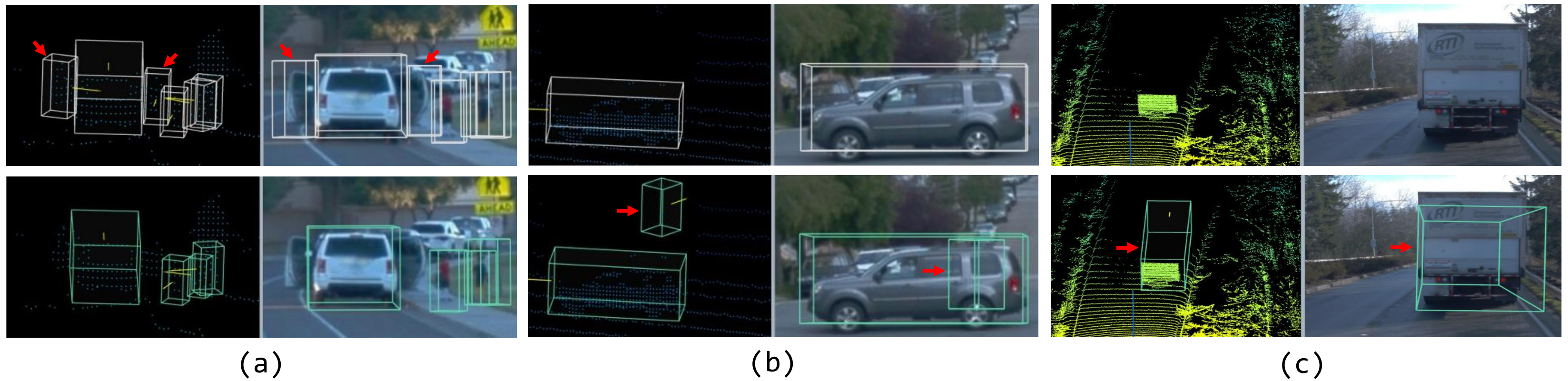}
    \caption{Qualitative analysis of the Waymo Open Dataset results. For each subgraph, we show a point cloud~(left), as well as the 3D bounding boxes projected into the image~(right) for clearer visualization. The first and second rows respectively show the results of the baseline {\dag}CenterPoint-Pillar and our method HS\&ST-Pillar. And the differences worth paying attention to be marked by red arrows.}
    \label{fig:exp}
\end{figure*}

\noindent\textbf{Effects of main contributions.}~We investigate the individual components of our proposed Fine-grained Pillar with extensive ablation experiments. Our method is consist of two major ways to improve the point cloud representation of pillar, which are Sub-pillar with Position Encoding and Tiny-pillar with DFSA module. 

As shown in $1^{st}$ and $2^{nd}$ rows of Table~\ref{tab:exp-4}, simply applying Sub-pillar can get $+2.08\%$ mAPH improvement on LEVEL 2 for average of three categories. It validates our argument that Sub-pillar can get better object features from uneven distribution point clouds in the height dimension. Moreover, $3^{rd}$ row of Table~\ref{tab:exp-4} shows that our Position Encoding can further improve the detection performance with $+1.35\%$ mean mAPH LEVEL 2 gain by keeping height position information of each sub-pillar in 2D pseudo image feature map.

Besides, as shown in $1^{st}$ and $4^{th}$ rows of Table~\ref{tab:exp-4}, simply applying Tiny-pillar and previous 2D CNN backbone in \cite{lang2019pointpillars} can get $+6.87\%$ mAPH improvement on LEVEL 2 for average of three categories. It validates our argument that Tiny-pillar can get more fine-grained features due to small grid. $5^{th}$ row of Table~\ref{tab:exp-4} shows that our Dense Feature branches can further improve the detection performance with $+3.30\%$ mean mAPH LEVEL 2 gain by applying most convolution blocks on dense small-scale feature map. Moreover, $6^{th}$ row of Table~\ref{tab:exp-4} shows that our Sparse Attention can also get $+0.24\%$ mean mAPH LEVEL 2 gain by refining dense features with sparse distribution of objects when upsampling to large-scale.

In general, our improved pillar with fine-grained feature in last row can improve the baseline in first row with $+10.91\%$ mean mAPH LEVEL 2 gain, especially the detection accuracy of small objects is significantly improved.

\begin{table}[ht]
    \centering
    \begin{tabular}{c|c|c|c|c}
        \hline
        $N_h$ & Veh. & Ped. &  Cyc. & Mean \\
        \hline
        1 & 60.32 & 49.22 & 51.58 & 53.71 \\
        2 & 60.35 & 51.79 & 53.92 & 55.35 \\
        4 & 61.09 & 52.87 & 54.70 & 56.22 \\
        6 & 61.35 & 53.50 & 56.58 & 57.14 \\
        8 & 61.17 & 54.42 & 56.68 & 57.42 \\
        \hline
    \end{tabular}
    \caption{Effects of the sub-part number $N_h$ in our HS-Pillar. We compare on $20\%$ Waymo validation by per-class and average 3D LEVEL 2 mAPH.}
    \label{tab:exp-5}
\end{table}

\noindent\textbf{Effects of number of sub-pillars.}~We propose a High-aware Sub-pillar to get fine-grained features by splitting each pillar into $N_h$ sub-pillars with position encoding in the height dimension, so the split number $N_h$ is an important hyper-parameter for our method. In Table \ref{tab:exp-5}, we explore the impact of different $N_h$ on the detection accuracy of each category. It can be seen that the detection accuracy of every category increases significantly with the increase of the sub-pillar number $N_h$, because the uneven distribution of point clouds in the height dimension is a phenomenon that exists in every category as shown in Figure \ref{fig:method-1}. However, when the number $N_h$ is further increased, the detection accuracy of vehicle decreases, which is due to few points in small sub-pillar make it difficult to extract effective features.

\begin{table}[ht]
    \centering
    \begin{tabular}{c|c|c|c|c|c}
        \hline
        $\{S_i\}_{1}^{n}$ & $\{N_i\}_{1}^{n}$ & Veh. & Ped. &  Cyc. & Mean \\
        \hline
        \multirow{2}*{$\{4,8\}$} & $\{2,4\}$ & 64.88 & 57.18 & 63.96 & 62.01 \\
        ~ & $\{3,5\}$ & 65.71 & 57.17 & 64.62 & 62.50 \\
        \hline
        \multirow{2}*{$\{2,4\}$} & $\{2,4\}$ & 65.26 & 57.83 & 64.17 & 62.42\\
        ~ & $\{3,5\}$ & 66.24 & 58.84 & 65.85 & 63.64 \\
        \hline
    \end{tabular}
    \caption{Effects of the downsampling scales $\{S_i\}_{1}^{n}$ and convolution blocks numbers $\{N_i\}_{1}^{n}$ in our DFSA module. We compare on $20\%$ Waymo validation by per-class and average 3D LEVEL 2 mAPH.}
    \label{tab:exp-6}
\end{table}

\noindent\textbf{Effects of settings in DFSA module.}~We propose a Dense Feature and Sparse Attention module to extract features efficiently based on the sparsity of Tiny-pillar. The downsampling scales $\{S_i\}_{1}^{n}$ and convolution blocks numbers $\{N_i\}_{1}^{n}$ of each dense feature branch are important hyper-parameters in our DFSA module, so we explore the impact of different settings on the detection accuracy of each category in the Table~\ref{tab:exp-6}. It can be seen that the detection results of each category maintain good robustness to different settings. Large-scale downsampling can achieve fast speed with a little accuracy drop cost, while more convolution blocks can improve performance due to larger receptive field.

\subsection{Qualitative Analysis}
We provide qualitative results in Figure~\ref{fig:exp} to show that how our method improve the detection results. As shown in Figure~\ref{fig:exp}~(a), the baseline model incorrectly recognizes the car doors as pedestrians because its coarse-grained feature of pillar from entire height area can not distinguish the 
similar distributions of car door and pedestrian. Meanwhile our method can avoid such false positive by more fine-grained feature of sub-pillars from different height areas. The Figure~\ref{fig:exp}~(b) shows when the object contains too few points due to occlusion, it can be correctly detected by our method with fine-grained feature of tiny-pillar but missed by baseline. It can also be seen from Figure~\ref{fig:exp}~(c) that our method can detect large objects better than baseline relying on larger receptive field from DFSA module.
%-------------------------

%-------------------------
\section{Conclusion}
In this paper, we present two novel modules to improve pillar with fine-grained feature, named height-aware sub-pillar and sparsity-based tiny-pillar, for accurate 3D object detection from point clouds. Our height-aware sub-pillar adopts a novel height-aware position encoding, which can capture point features from different height areas to distinguish multi sub-pillars. Our sparsity-based tiny-pillar can extract detailed point features from small grids with our novel sparsity-based CNN backbone, which learns the position information of objects from large-scale feature maps and shape information of objects from small-scale feature maps efficiently by our DFSA module. Sufficient experiments show that both of our proposed two improvements significantly outperform the baseline model, and the improved pillar model can achieve new state-of-the-art performance on the Waymo Open Dataset.
%-------------------------

\bibliography{ref.bib}

\end{document}